\documentclass[11pt,a4paper]{article}
\usepackage[hyperref]{acl2021}
\usepackage{times}
\usepackage{multirow}
\usepackage{multicol}
\usepackage{latexsym}

\usepackage{booktabs} 
\usepackage[normalem]{ulem}
\usepackage{xcolor} 
\usepackage{algorithm}
\usepackage[noend]{algpseudocode}
\usepackage{amsmath}
\usepackage{mathrsfs}
 \usepackage{amssymb}
\usepackage{subfigure}
\usepackage{makecell}
\usepackage{mathtools}
\usepackage[font=rm]{caption}
\DeclareCaptionType{copyrightbox}
\usepackage{shortvrb}
\usepackage{tabularx}
\usepackage{verbatim}
\usepackage{xspace}
\usepackage{listings}
\usepackage{tikz}
\usepackage{verbatim}
\usetikzlibrary{trees}
\usepackage{forest}
\forestset{
    nice empty nodes/.style={
        for tree={
            s sep=0.1em, 
            l sep=0.33em,
            inner ysep=0.4em, 
            inner xsep=0.05em,
            l=0,
            calign=midpoint,
            fit=tight,
            where n children=0{
               tier=word,
               minimum height=1.25em,
            }{},
            where n children=2{
               l-=1em,
            }{},
            parent anchor=south,
            child anchor=north,
            delay={if content={}{
                    inner sep=0pt,
                    edge path={\noexpand\path [\forestoption{edge}] 
                    			(!u.parent anchor) 
                               -- (.south)\forestoption{edge label};}
                }{}}
        },
    },
}
\usepackage{fontawesome}
\usepackage[multiple]{footmisc}
\usepackage[all]{nowidow}
\usepackage{balance}
\usepackage{microtype}

\usepackage{xspace}
\newcommand{\bert}{\textsc{Bert}\xspace}
\newcommand{\gumbel}{\textsc{Gumbel}\xspace}

\newcommand{\pruning}{\textsc{Pruning}\xspace}
\newcommand{\find}{\textsc{Find}\xspace}
\newcommand{\growing}{\textsc{TreeInduction}\xspace}
\newcommand{\argmax}{\mathop{\mathrm{argmax}}\nolimits}

\aclfinalcopy 
\def\aclpaperid{326} 

\usepackage{gnuplottex}
\usepackage[medium,compact]{titlesec}
\usepackage{enumitem}
\setlist{itemsep=0pt,parsep=0pt}

\newcommand{\Patk}[1]{\mbox{\Pat@$#1$}}
\newcommand{\RBPatp}[1]{\mbox{\RBP@$#1$}}

\newcommand{\NDCGatk}[1]{\mbox{\NDCG@$#1$}}
\newcommand{\ERRatk}[1]{\mbox{\ERR@$#1$}}

\newcommand{\myurl}[1]{{\url{#1}}}

\newcommand{\mycomment}[1]{}

\newlength{\onedigit}
\newcounter{todocount}
\title{R2D2: Recursive Transformer based on Differentiable Tree\\
for Interpretable Hierarchical Language Modeling}

\author{Xiang Hu\footnotemark[2] \thanks{~~Equal contribution.} \quad Haitao Mi\footnotemark[2] \footnotemark[1] \quad Zujie Wen\footnotemark[2] \quad Yafang Wang\footnotemark[2] \\
  \bf Yi Su\footnotemark[2] \quad Jing Zheng\footnotemark[2] \quad Gerard de Melo\footnotemark[3] \\
  Ant Financial Services Group\footnotemark[2] \\
  \tt \{aaron.hx, haitao.mi, zujie.wzj, yafang.wyf, yi.su, jing.zheng\} \\
  \tt @alibaba-inc.com\footnotemark[2] \\
  Hasso Plattner Institute / University of Potsdam\footnotemark[3] \\
  \tt gdm@demelo.org\footnotemark[3]  \\}

\date{}

\begin{document}
\maketitle
\begin{abstract}

Human language understanding operates at multiple levels of granularity (e.g., words, phrases, and sentences) with
increasing levels of abstraction that can be hierarchically combined.
However, existing deep models with stacked layers do not explicitly model any sort of hierarchical process. 
This paper proposes a recursive Transformer model based on differentiable CKY style binary trees to emulate the composition process. 
We extend the bidirectional language model pre-training objective to this architecture, attempting to predict each word given its left and right abstraction nodes.
To scale up our approach, we also introduce an efficient pruned tree induction algorithm to enable encoding in just a linear number of composition steps.
Experimental results on language modeling and unsupervised parsing show the effectiveness of 
our approach.\footnote{The code is available at: \url{https://github.com/alipay/StructuredLM\_RTDT}}
\end{abstract}

\section{Introduction}
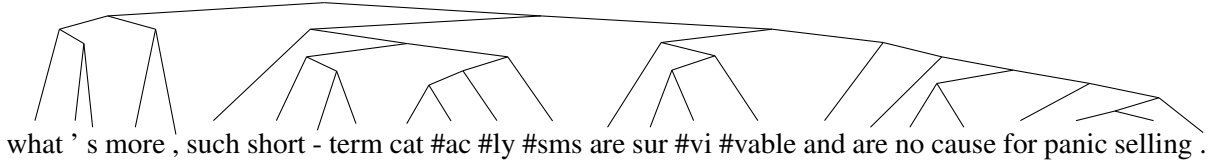
\begin{figure*}[!htb]
\begin{forest} 
	shape=coordinate,
	where n children=0{
		tier=word
	}{},
	nice empty nodes
    [[[[what][['][s]]][[more][{,}]]][[[such][[[short][[-][term]]][[[[cat][\#ac]][\#ly]][\#sms]]]][[[are][[[sur][\#vi]][\#vable]]][[and][[are][[[no][cause]][[for][[[panic][selling]][\\.]]]]]]]]] 
\end{forest}
\caption{An example output tree emerging from our proposed method.}
\label{fig:tree_example}
\end{figure*}

The idea of devising a structural model of language capable of learning 
both representations and meaningful syntactic structure without any human-annotated trees has been a long-standing but challenging goal.
Across a diverse range of linguistic theories, human language is assumed to possess a recursive hierarchical structure~\cite{DBLP:journals/tit/Chomsky56,chomsky2014aspects,de-marneffe-etal-2006-generating} 
such that lower-level meaning is combined to infer higher-level semantics.
Humans possess notions of characters, words, phrases, and sentences, which children naturally learn to segment and combine.

Pretrained language models such as BERT~\cite{DBLP:conf/naacl/DevlinCLT19} have achieved substantial gains across a range of tasks. 
However, they simply apply layer-stacking with a fixed depth to increase the modeling power~\cite{bengio2009learning,lecun2015deep}. Moreover,
as the core Transformer component~\cite{vaswani2017attention} does not capture positional information, 
one also needs to incorporate additional positional embeddings.
Thus, pretrained language models do not explicitly reflect the hierarchical structure of linguistic understanding.

Inspired by \newcite{le-zuidema-2015-forest}, \newcite{DBLP:journals/corr/MaillardCY17} proposed a fully differentiable CKY parser to model the hierarchical process explicitly. 
To make their parser differentiable, they primarily introduce an energy function to combine all possible derivations when constructing each cell representation.
However, their model is based on Tree-LSTMs~\cite{tai2015improved,zhu2015long} and requires $O(n^3)$ time complexity. Hence, it is hard to scale up to large training data.

In this paper, we revisit these ideas, and propose a model applying recursive Transformers along differentiable trees (R2D2).
To obtain differentiability, we adopt Gumbel-Softmax estimation \cite{DBLP:conf/iclr/JangGP17} as an elegant solution. Our encoder parser operates in a bottom-up fashion akin to CKY parsing, yet runs in linear time with regard to the number of composition steps, thanks to a novel pruned tree induction algorithm. 
As a training objective, the model seeks to recover each word in a sentence given its left and right syntax nodes.
Thus, our model does not require any positional embedding and does not need to mask any words during training. Figure~\ref{fig:tree_example} presents an example binary tree induced by our method: Without any syntactic supervision, it acquires a model of hierarchical construction from the word-piece level to words, phrases, and finally the sentence level.

We make the following contributions:
\begin{itemize}[leftmargin=*,noitemsep,nolistsep]
    \item Our novel CKY-based recursive Transformer on differentiable trees model is able to learn 
          both representations and tree structure (Section~\ref{sec:model}).
    \item We propose an efficient 
    optimization algorithm to scale up our approach to a linear number of composition steps (Section~\ref{sec:pruning_algo}).
    \item We design an effective pre-training objective, which predicts each word given its left and right syntactic nodes (Section~\ref{sec:train_obj}).
\end{itemize}
For simplicity and efficiency reasons, in this paper we conduct experiments only on the tasks of language modeling and
unsupervised tree induction. 
The experimental results on language modeling show that our model significantly outperforms baseline models with same parameter size even in fewer training epochs. 
At unsupervised parsing, our model as well obtains competitive results.

\section{Methodology}

\subsection{Model Architecture}\label{sec:model}

\begin{figure}[htb!]
  \centering
  \includegraphics[width=0.45\textwidth]{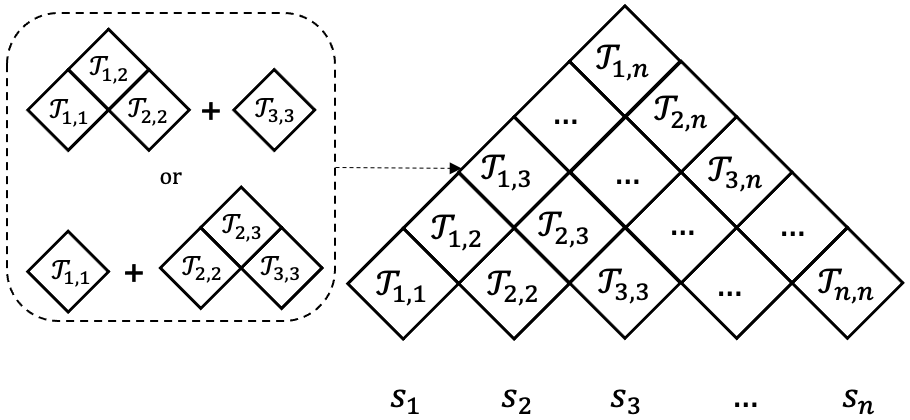}
  \caption{Chart data structure. There are two alternative ways of generating $\mathcal{T}_{1,3}$:
     combining either ($\mathcal{T}_{1,2}$, $\mathcal{T}_{3,3}$) or ($\mathcal{T}_{1,1}$, $\mathcal{T}_{2,3}$).}
  \label{fig:chart_data}
\end{figure}

\paragraph{Differentiable Tree.} 
We follow \newcite{DBLP:journals/corr/MaillardCY17} in defining a differentiable binary parser using
a CKY-style~\cite{10.5555/1097042,kasami1966efficient,younger1967recognition} encoder.
Informally, given a sentence $\mathbf{S} = \{s_{1}, s_{2},..., s_{n}\}$ with $n$ words or word-pieces, 
Figure~\ref{fig:chart_data} shows the chart data structure $\mathcal{T}$, 
where each cell $\mathcal{T}_{i, j}$ is a tuple $\langle e_{i, j}, p_{i, j}, \widetilde{p}_{i,j} \rangle$, 
$e_{i, j}$ is a vector representation, $p_{i, j}$ is the probability of a single composition step, 
and $\widetilde{p}_{i,j}$ is the probability of the subtree at span $[i, j]$ over sub-string $s_{i:j}$.
At the lowest level, we have terminal nodes $\mathcal{T}_{i, i}$ with $e_{i, i}$ initialized as embeddings of inputs $s_{i}$, while
$p_{i, i}$ and $\widetilde{p}_{i,i}$ are set to one. 
When $j>i$, the representation $e_{i, j}$ is a weighted sum of intermediate combinations $c_{i, j}^{k}$, defined as: 
\begin{align}
\label{eq:tree_encoder}
&c_{i,j}^{k}, \  p_{i,j}^{k} = f(e_{i, k}, e_{k+1, j})\\
&\widetilde{p}_{i,j}^{k} = p_{i,j}^{k} \,\, \widetilde{p}_{i,k} \,\, \widetilde{p}_{k+1,j}\\
&\boldsymbol{\alpha}_{i,j} = \text{\gumbel} (\log( \mathbf{\widetilde{p}}_{i,j}))\\
&e_{i,j} =  [c_{i,j}^{i}, c_{i,j}^{i+1}, ..., c_{i,j}^{j-1}]\boldsymbol{\alpha}_{i,j}\\
&[p_{i,j},\widetilde{p}_{i,j}] = \boldsymbol{\alpha}_{i,j}^\intercal [\boldsymbol{p}_{i,j}, \boldsymbol{\widetilde{p}}_{i,j}]
\end{align}
Here, $k$ is a split point from $i$ to $j-1$, $f(\cdot)$ is a composition function that we shall further define later on,
$p_{i,j}^{k}$ and $\widetilde{p}_{i,j}^{k}$ denote the single step combination probability and the subtree probability, respectively, at split point $k$,
$\boldsymbol{p}_{i,j}$ and $\boldsymbol{\widetilde{p}}_{i,j}$ are the concatenation of all $p_{i,j}^{k}$ or $\widetilde{p}_{i,j}^{k}$ values, 
and \gumbel is the Straight-Through Gumbel-Softmax operation of \newcite{DBLP:conf/iclr/JangGP17} with temperature set to one. The $[,]$ notation denotes stacking of tensors.

\begin{figure}[!htb]
    \centering
    \includegraphics[width=0.4\textwidth]{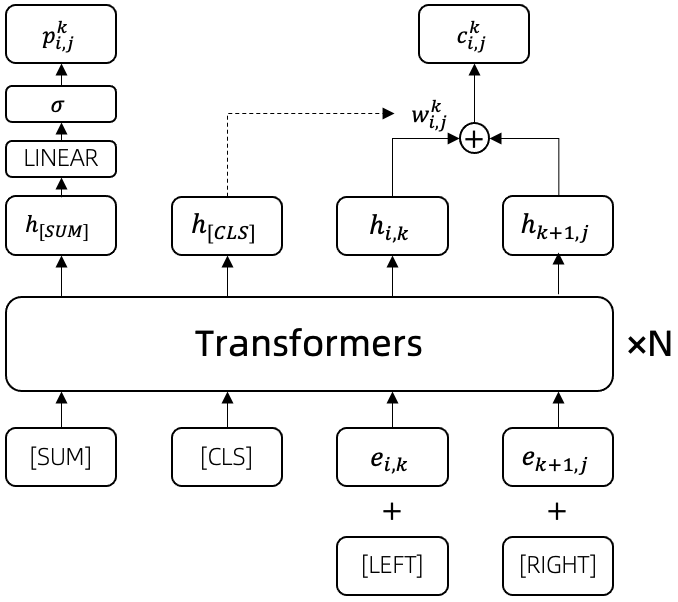}
    \caption{Recursive Transformer-based encoder.}
    \label{fig:tree_encoder}
\end{figure}

\paragraph{Recursive Transformer.} 
Figure \ref{fig:tree_encoder} provides a schematic overview of the composition function $f(\cdot)$, comprising $N$ Transformer layers. 
Taking $c_{i,j}^{k}$ and $p_{i,j}^{k}$ as an example, the input is a concatenation of two special tokens \texttt{[SUM]} and \texttt{[CLS]}, 
the left cell $e_{i, k}$, and the right cell $e_{k+1, j}$. We also add role embeddings 
(\texttt{[LEFT]} and \texttt{[RIGHT]}) to the left and right inputs, respectively. 
Thus, the input consists of four vectors in $\mathbb{R}^d$.
We denote as $h_{\texttt{[SUM]}}, h_{\texttt{[CLS]}}, h_{i, k}, h_{k+1, j} \in \mathbb{R}^{d}$ the hidden state of 
the output of $N$ Transformer layers. 
This is followed by a linear layer over $h_{\texttt{[SUM]}}$ to obtain
\begin{equation}
\label{eq:comp_p}
p_{i,j}^{k} = \sigma (W_{p} h_{\texttt{[SUM]}} + b_{p}),
\end{equation}
where $W_{p}\in \mathbb{R}^{1 \times d}$, $b_{p}\in \mathbb{R}$, and $\sigma$ refers to sigmoid activation.
Then, $c_{i,j}^{k}$ is computed as
\begin{equation}
\label{eq:tree_encoder_detail}
\begin{aligned}
&w_{i,j}^{k} = \mathrm{softmax}(W_{w} h_{\texttt{[CLS]}} + b_{w}) \\
&c_{i,j}^{k}= [h_{i, k}, h_{k+1, j}] w_{i,j}^{k},
\end{aligned}
\end{equation}
where $W_{w} \in \mathbb{R}^{2 \times d}$ with
$w_{i,j}^{k} \in \mathbb{R}^{2}$ capturing the respective weights of the left and right hidden states $h_{i, k}$ and $h_{k+1, j}$,
and the final $c_{i,j}^{k}$ is a weighted sum of $h_{i, k}$ and $h_{k+1, j}$.

\begin{figure*}
    \flushleft
    \includegraphics[width=1\textwidth]{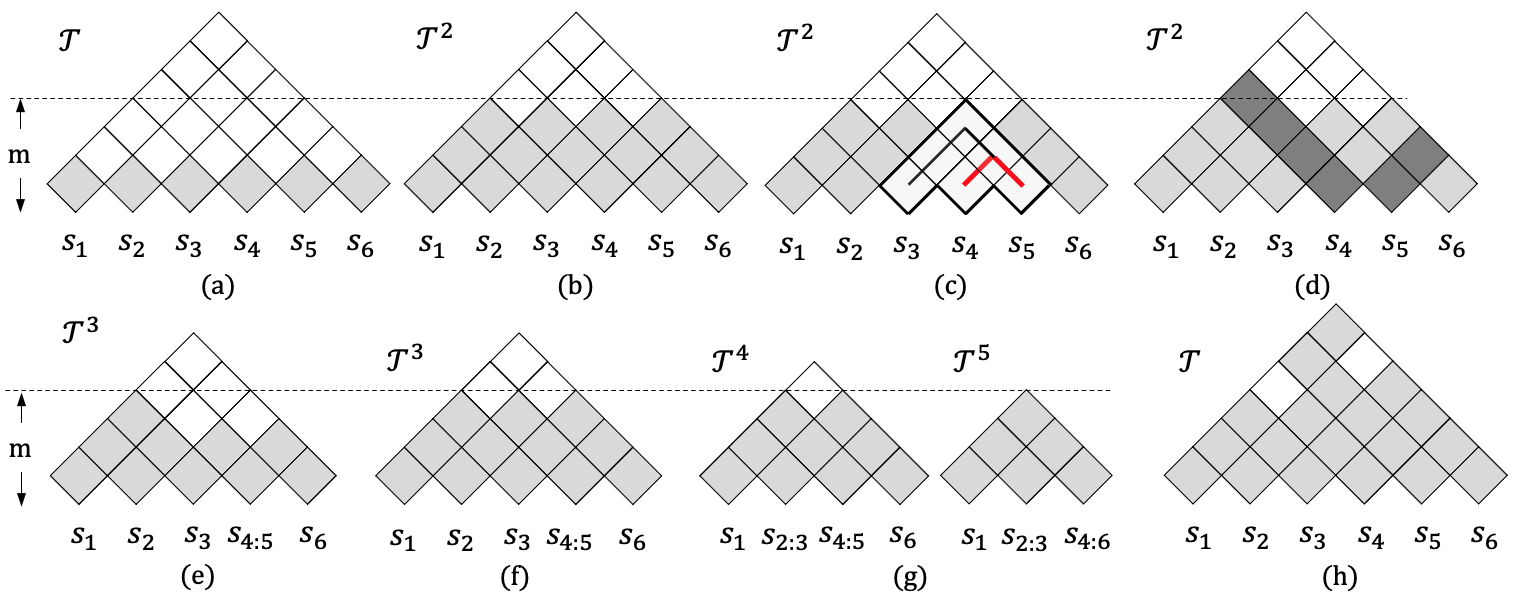}
    \caption{Example of encoding. (a) Initialized chart table.
    (b) Row-by-row encoding up to pruning threshold $m$. 
    (c) For each cell in the $m$-{th} row, recover its subtree and collect candidate nodes, each of which must appear in the subtree and also must be in the $2$nd row, 
        e.g., the tree of $\mathcal{T}^{2}_{3,5}$ is within the dark line, and the candidate node is $\mathcal{T}^{2}_{4,5}$.
    (d) Find locally optimal node, which is $\mathcal{T}^{2}_{4,5}$ here, and treat span $s_{4:5}$ as non-splittable. Thus, 
        the dark gray cells become prunable. 
    (e) Construct a new chart table $\mathcal{T}^3$ treating cell $\mathcal{T}^{2}_{4,5}$ as a new \emph{terminal} node and eliminating the prunable cells.
    (f) Compute empty cells in $m$-{th} row. 
    (g) Keep pruning and growing the tree until no further empty cells remain. 
    (h) Final discrete chart table.}
    \label{fig:prune_step}
\end{figure*}

\paragraph{Tree Recovery.}
As the Straight-Through Gumbel-Softmax picks the optimal splitting point $k$ at each cell in practice,
it is straightforward to recover the complete derivation tree, \texttt{Tree}($\mathcal{T}_{1,n}$), from the root node $\mathcal{T}_{1,n}$ in a top-down manner recursively.

\subsection{Complexity Optimization}\label{sec:pruning_algo}

\begin{algorithm}[!h]
\small
    \caption{\label{alg:prune_grow} Pruned Tree Induction Algorithm}
    \begin{algorithmic}[1]
        \Require $\mathcal{T}=$ 2-d array holding cell references
        \Require $m=$ pruning threshold
        \Function {Pruning}{$\mathcal{T}$, $m$}
            \State{$u \gets$ \find($\mathcal{T}$, $m$)} \Comment{Find optimal merge point}
            \State{$n \gets \mathcal{T}.len$}
            \State{$\mathcal{T}' \gets [n-1][n-1]$}\Comment{Create a new 2-d array}\label{alg:t_prime_start}
            \For{$i$ $\in$ $1$ to $n-1$}
                \For{$j$ $\in$ $i$ to $n-1$}
                    \State{$i' \gets i \geq u+1$ ~?~ $i+1$ ~:~ $i$}
                    \State{$j' \gets j \geq u$ ~?~ $j+1$ ~:~ $j$}
                    \State{$\mathcal{T}'_{i,j} \gets \mathcal{T}_{i',j'}$} \Comment{Skip dark gray cells in Fig.~\ref{fig:prune_step}}
                \EndFor
            \EndFor \label{alg:t_prime_end}
            \State{\Return{$\mathcal{T}'$}}
        \EndFunction
        
        \Function {TreeInduction}{$\mathcal{T}$, $m$}
            \State{$\mathcal{T}' \gets \mathcal{T}$}
            \For{$t$ $\in$ $1$ to $\mathcal{T}.len-1$ \label{alg:continue_grow}}
                \If{$t \geq m$}
                    \State{$\mathcal{T}' \gets $ \pruning($\mathcal{T}'$,$m$)} \label{alg:call_prune}
                \EndIf
                \State{$l \gets $ min($t+1$, $m$)}\Comment{Clamp the span length}
                \For{$i$ $\in$ $1$ to $\mathcal{T}'.len-l+1$}
                    \State{$j \gets i+l-1$}
                    \If{$\mathcal{T}'_{i,j}$ is empty}
                        \State{Compute cell $\mathcal{T}'_{i,j}$ with Equation~\ref{eq:tree_encoder}}
                    \EndIf
                \EndFor
            \EndFor
            \State{\Return{$\mathcal{T}$}}
        \EndFunction
    \end{algorithmic}
\end{algorithm}

As the core computation comes from the composition function $f(\cdot)$, our {\it pruned tree induction algorithm} aims to 
reduce the number of composition calls from $O(n^{3})$ in the original CKY algorithm to linear.

Our intuition is based on the conjecture that locally optimal compositions
are likely to be retained and participate in higher-level feature combination.
Specifically, taking $\mathcal{T}^2$ in Figure~\ref{fig:prune_step} (c) as an example, 
we only pick locally optimal nodes from the second row of $\mathcal{T}^2$. 
If $\mathcal{T}_{4,5}^2$ is locally optimal and non-splittable, 
then all the cells highlighted in dark gray in (d) may be pruned, as they break span $[4, 5]$.
For any later encoding, including higher-level ones, we can merge the nodes and treat $\mathcal{T}_{4,5}^2$ as a new non-splittable terminal node (see (e) to (g)).

\begin{algorithm}[!h]
\small
    \caption{\label{alg:find_best_merge} Find the best merge point}
    \begin{algorithmic}[1]
        \Require $\mathcal{T}=$ 2-d array holding cell references
        \Require $m=$ pruning threshold
        \Function {Find}{$\mathcal{T}$, $m$}
            \State{$n \gets \mathcal{T}.len$}
            \State{$\mathcal{L} \gets [n-1]$}\Comment{Create an array}
            \For{$i$ $\in$ $1$ to $n-1$}
                \State{$\mathcal{L}[i] \gets \mathcal{T}_{i,i+1}$}\Comment{Collect cells on the 2{nd} row}
            \EndFor
            \State{$\tau \gets \emptyset$\label{alg:bn_start}} 
            \For{$i$ $\in$ $1$ to $n-m+1$}\Comment{Iterate to $m$-th row}
                \State{$j = i+m-1$}
                \State{$\tau \gets \tau \cup \{ c \mid c \in \texttt{Tree}(\mathcal{T}_{i,j}) \wedge c \in \mathcal{L} \}$ } 
            \EndFor\label{alg:bn_end}
            \State{$l \gets \mathbf{new}\ \mathrm{List()}$}
            \For{cell $x$ $\in$ $\tau$} \label{alg:find_s}
                \State{$i \gets \mathcal{L}.$index($x$)}
                \State{$\overline{p}_{l} \gets 1-\mathcal{L}[i-1].p$\label{alg:left_p}}
                \State{$\overline{p}_{r} \gets 1-\mathcal{L}[i+1].p$\label{alg:right_p}}\\
                \Comment{{If index out of boundary then set to 0}}
                \State{$l$.append($x.p \cdot \overline{p}_{l} \cdot \overline{p}_{r}$)}
            \EndFor 
            \State{\Return{$\argmax_{i}{l[i]}$} \label{alg:find_e}}
        \EndFunction
    \end{algorithmic}
\end{algorithm}

Figure~\ref{fig:prune_step} walks through the steps of processing a sentence of length 6, where $s_{i:j}$ denotes a sub-string from $s_i$ to $s_j$.
Algorithm~\ref{alg:prune_grow} constructs our chart table $\mathcal{T}$ sequentially row-by-row. 
Let $t$ be the time step and $m$ be the pruning threshold.
First, we invoke \growing$(\mathcal{T}, m)$, 
and compute a row of cells at each time step when $t < m$ as in regular CKY parsing, leading to result (b) in Figure~\ref{fig:prune_step}.
When $t \geq m$, we call \pruning$(\mathcal{T}, m)$ in Line~\ref{alg:call_prune}.
As mentioned, the \pruning function aims to find the locally optimal 
combination node in $\mathcal{T}$, prunes some cells, and returns a new table omitting the pruned cells.
Algorithm~\ref{alg:find_best_merge} shows how we \find the locally optimal combination node.
Again, the candidate set for the locally optimal node is the second row of $\mathcal{T}$,
and we also take advantage of the subtrees derived from all nodes in the $m$-{th} row to limit the 
candidate set. Lines \ref{alg:bn_start} to \ref{alg:bn_end} in Algorithm~\ref{alg:find_best_merge}
generate the candidate set. Each candidate must be in the second row of $\mathcal{T}$ and also 
must be used in a subtree of any node in the $m$-th row.
Given the candidate set, we find the least ambiguous one as the optimal selection 
(Lines \ref{alg:find_s} to \ref{alg:find_e}), 
i.e., the node with maximum own probability while adjacent bi-gram node probabilities (Lines \ref{alg:left_p} and \ref{alg:right_p} ) are as low as possible. 
After selecting the best merge point $u$, 
cells in $\{\mathcal{T}^{t}_{i,j} \mid j=u\}\cup\{\mathcal{T}^{t}_{i,j} \mid i=u+1\}$ are pruned (highlighted in dark gray in (d)),
and we generate a new table $\mathcal{T}^{t+1}$ by removing pruned nodes 
(Lines \ref{alg:t_prime_start} to \ref{alg:t_prime_end} in Algorithm~\ref{alg:prune_grow}). 
Then we obtain (e), and compute the empty cells on the $m$-{th} row of $\mathcal{T}^{3}$ to obtain (f). 
We continue with the loop in Line \ref{alg:continue_grow}, 
trigger \pruning again, and obtain a new table $\mathcal{T}^{t+1}$, and then fill empty cells on the $m$-th row $\mathcal{T}^{t+1}$.
Continuing with the process until all cells are computed, as shown in (g), 
we finally obtain a discrete chart table as given in (h).

In terms of the time complexity, when $t \geq m$, there are at most $m$ cells to update, so the complexity of each step is less than $O(m^{2})$. 
When $t \leq m$, the complexity is $O(t^{3}) \leq O(m^{2}t)$.
Thus, the overall times to call the composition function is $O(m^{2}n)$, which is linear considering $m$ is a constant.

\subsection{Pretraining}\label{sec:train_obj}

Different from the masked language model training of \bert, we directly minimize the sum of all negative log probabilities of all 
words or word-pieces $s_{i}$ given their left and right contexts.
\begin{equation}
\underset{\theta}{\mathrm{min}}\,\sum_{i=1}^{n} -\log\,p_{\theta}(s_{i} \mid s_{1:i-1}, s_{i+1:n})
\end{equation}

\begin{figure}
    \centering
    \includegraphics[width=0.45\textwidth]{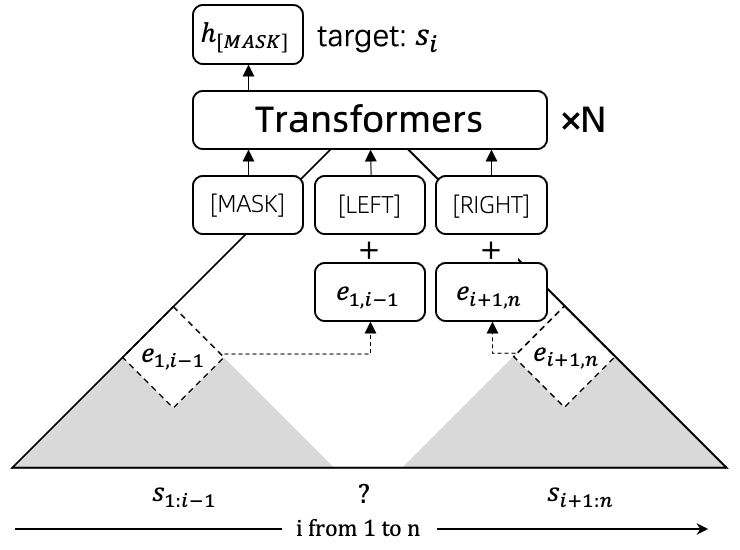}
    \caption{The objective for our pretrained model.}
    \label{fig:language_model_task}
\end{figure}

As shown in Figure~\ref{fig:language_model_task}, 
after invoking our recursive encoder on a sentence $\mathbf{S}$, 
we directly use $e_{1,i-1}$ and $e_{i+1,n}$ as the left and right contexts, respectively, for each word $s_i$.
To distinguish from the encoding task, the input consists of a concatenation of a special token \texttt{[MASK]}, $e_{1,i-1}$, and $e_{i+1,n}$.
We apply the same composition function $f(\cdot)$ as in Figure~\ref{fig:tree_encoder}, 
and feed $h_{\texttt{[MASK]}}$ through an output softmax to predict the distribution of $s_i$ over the complete vocabulary.
Finally, we compute the cross-entropy over the prediction and ground truth distributions.

In cases where $e_{1,i-1}$ or $e_{i+1,n}$ is missing due to the pruning algorithm in Section~\ref{sec:pruning_algo}, 
we simply use the left or right longest adjacent non-empty cell. 
For example, $\mathcal{T}_{x, i-1}$ means the longest non-empty cell assuming we cannot find any non-empty $\mathcal{T}_{x', i-1}$ for all $x'< x$.
Analogously, $\mathcal{T}_{i+1, y}$ is defined as the longest non-empty right cell.
Note that although the final table is sparse, the sentence representation $e_{1,n}$ is always established.

\section{Experiments}
As our approach (R2D2) is able to learn both representations and intermediate structure,
we evaluate its representation learning ability on bidirectional language modeling and evaluate the intermediate structures on unsupervised parsing.

\subsection{Bidirectional Language Modeling}

\subsubsection{Setup}\label{sec:LM_setup}

\paragraph{Baselines and Evaluation.} 
As the objective of our model is to predict each word with its left and right context, 
we use the \emph{pseudo-perplexity} (PPPL) metric of \newcite{salazar-etal-2020-masked} to evaluate bidirectional language modeling.
\begin{align*}
\mathcal{L}(\mathbf{S})&=\frac{1}{n} \sum_{i=1}^{n} \mathrm{log}P(s_{i} \mid s_{1:i-1}, s_{i+1:n}, \theta)\\
\text{PPPL}(\mathbf{S})&=\mathrm{exp}\left(-\frac{1}{N}\sum_{j=1}^{N}\mathcal{L}(\mathbf{S}^{j})\right)
\end{align*}
PPPL is a bidirectional version of perplexity, establishing a macroscopic assessment of the model’s ability to deal with diverse linguistic phenomena. 

We compared our approach with SOTA autoencoding and autoregressive language models capable of capturing bidirectional contexts, including \bert, XLNet~\cite{yang2019xlnet}, and ALBERT~\cite{lan2019albert}. 
For a fair apples to apples comparison, all models use the same vocabulary and are trained from scratch on a language modeling corpus. 
The models are all based on the open source Transformers library\footnote{https://github.com/huggingface/transformers}. 
To compute PPPL for models based on sequential Transformers, 
for each word $s_{i}$, we only mask $s_{i}$ while others remain visible to predict $s_{i}$. 
When we evaluate our R2D2 model, for each word $s_{i}$, we treat the left $s_{1:i-1}$ and right $s_{i+1:n}$ as two complete sentences separately, 
then encode them separately, and pick the root nodes as the final representations of left and right contexts. In the end, we predict word $s_i$ by running our Transformers as in Figure~\ref{fig:language_model_task}.

\paragraph{Data.} The English language WikiText-2 corpus~\cite{DBLP:conf/iclr/MerityX0S17} serves as training data. 
The dataset is split at the sentence level, and sentences longer than 128 after tokenization are discarded (about 0.03\% of the original data). 
The total number of sentences is 68,634, and the average sentence length is 33.4.

\paragraph{Hyperparameters.}  The tree encoder of our model uses 3-layer Transformers with 768-dimensional embeddings, 
3,072-dimensional hidden layer representations, and 12 attention heads. Other models based on the Transformer share the same setting but vary on the number of layers.
Training is conducted using Adam optimization with weight decay with a learning rate of $5 \times 10^{-5}$.
The batch size is set to 8 for $m$=$8$ and 32 for $m$=$4$, though we also limit the maximum total length for each batch, such that excess sentences are moved to the next batch. The limit is set to 128 for $m$=$8$ and 512 for $m$=$4$. It takes about 43 hours for 10 epochs of training with $m=8$ and about 9 hours with $m$=$4$, on 8 v100 GPUs.

\subsubsection{Results and Discussion}

\begin{table}
\begin{center}
\setlength{\tabcolsep}{2pt}
\resizebox{0.48\textwidth}{!}{
\begin{tabular}{@{}c|r r c l r c@{}}
           & \#param & \#layer & \#epoch & cplx & PPPL  \\ \hline \hline
\bert    & 46M  & 3 &10&$O(n^{2})$& 441.42 \\
XLNet   & 46M & 3 &10 &$O(n)$& 301.87 \\
ALBERT  & 46M & 12 & 10 & $O(n^{2})$ & 219.20 \\
XLNet   & 116M& 12&10 &$O(n)$& 127.74 \\
\bert   & 109M& 12&10 &$O(n^{2})$& 103.54 \\ \hline
T-LSTM ($m$=4) & 46M & 1 &10 &$O(n)$& 820.57 \\
Ours ($m$=4)  & 45M & 3 &10 &$O(n)$& 83.10    \\
Ours ($m$=8)  & 45M & 3 &10 &$O(n)$& \textbf{57.40} \\
\hline \hline
\bert       & 46M& 3 & 60 &$O(n^{2})$& 112.17          \\
XLNet      & 46M& 3 & 60 &$O(n)$&  105.64        \\
ALBERT     & 46M& 12& 60 & $O(n^{2})$ & 71.52 \\
XLNet      & 116M& 12 &60 & $O(n)$& 59.74        \\
\bert       & 109M& 12 &60 &$O(n^{2})$&\textbf{44.70}          \\ \hline
Ours ($m$=4)  & 45M& 3 &60 &$O(n)$& 55.70         \\
Ours ($m$=8)  & 45M& 3 &60 &$O(n)$& 54.60          \\ \hline
\end{tabular}
}
\end{center}
\caption{Comparison with state-of-the-art models trained from scratch on WikiText-2 with different settings (number of Transformer layers and training epochs). $m$ is the pruning threshold.\label{tab:blm}}
\end{table}

Table~\ref{tab:blm} presents the results of all models with different parameters.
Our model outperforms other models with the same parameter size and number of training epochs.
These results suggest that our model architecture utilizes the training data more efficiently.
Comparing the different pruning thresholds $m$=$4$ and $m$=$8$ (last two rows), 
the two models actually converge to a similar place after 60 epochs, 
confirming the effectiveness of the pruned tree induction algorithm.
We also replace Transformers with Tree-LSTMs as in \newcite{DBLP:conf/iclr/JangGP17}, denoted as T-LSTM, finding that the perplexity is significantly higher compared to other models.

The best score is from the \bert model with 12 layers at epoch 60.
Although our model has a linear time complexity, it is still a sequential encoding model, 
and hence its training time is not comparable to that of fully parallelizable models.
Thus, we do not have results of 12-layer Transformers in Table~\ref{tab:blm}. 
The experimental results comparing models with the same parameter size suggest that our model may perform even better with further deep layers.

\begin{table}[]
\centering
\resizebox{0.48\textwidth}{!}{
\begin{tabular}{l|c|c}
& {\it emb.} $\times$ {\it hid.} $\times$ {\it lay.}  & training time \\ \hline \hline
Ours (m=4)         & $768 \times 3072 \times 3$ & 7h          \\
~~-w/o pruning    & $12 \times 12 \times 1$    & 1125h       \\ 
~~-w/o pruning$^*$ & $768 \times 3072 \times 3$ & $5 \times 10^7$h \\ \hline
\end{tabular}
}
\caption{Training time for one epoch on a single v100 GPU, where {\it emb.} and {\it hid.} represent the dimensions of word embeddings and hidden state respectively, and {\it lay.} is the number of transformer layers.
$^*$ means proportionally estimated time.}\label{tab:acc}
\end{table}

Table~\ref{tab:acc} shows the training time of our R2D2 with and without pruning.
The last row is proportionally estimated by running the small setting ($12 \times 12 \times 1$).
It is clear that it is not feasible to run our R2D2 without pruning.

\subsection{Unsupervised Constituency Parsing}
We next assess to what extent the trees that naturally arise in our model bear similarities with human-specified parse trees.

\subsubsection{Setup}
\paragraph{Baselines and Evaluation.} 
For comparison, we further include four recent strong models for unsupervised parsing with open source code: \bert masking \cite{DBLP:conf/acl/WuCKL20}, Ordered Neurons \cite{DBLP:conf/iclr/ShenTSC19}, DIORA \cite{DBLP:conf/naacl/DrozdovVYIM19} and C-PCFG \cite{kim-etal-2019-compound}. 
Following~\newcite{htut-etal-2018-grammar}, we train all systems on a training set consisting of raw text, and evaluate and report the results on an annotated test set. 
As an evaluation metric, we adopt sentence-level unlabeled $F_1$ computed using the script from \newcite{kim-etal-2019-compound}.
We compare against the non-binarized gold trees per convention.
The best checkpoint for each system is picked based on scores on the validation set. 

As our model is a pretrained model based on word-pieces, 
for a fair comparison, we test all models with two types of input: word level (W) and word-piece level (WP)\footnote{As DIORA relies on ELMO word embeddings, 
to support word-piece level inputs, we use BERT word-piece embeddings instead.}. 
To support word-piece level evaluation, 
we convert gold trees to word-piece level trees 
by simply breaking each terminal node into a non-terminal node with its word-pieces as terminals, e.g., (NN discrepancy) into (NN (WP disc) (WP \#\#re) (WP \#\#pan) (WP \#\#cy).
We set the pruning threshold $m$ to 8 for our tree encoder.

To support a word-level evaluation, since our model uses word-pieces, 
we force it to not prune or select spans that conflict with word spans during prediction, and then merge word-pieces into words in the final output. However, note that this constraint is only used for word-level prediction.

For training, we use the same hyperparameters as in Section~\ref{sec:LM_setup}. Our model pretrained on WikiText-2 is finetuned on the training set with the same unsupervised loss objective. For Chinese, we use a subset of Chinese Wikipedia for pretraining, specifically the first 100,000 sentences shorter than 150 characters.

\paragraph{Data.}  We test our approach on the Penn Treebank (PTB) \cite{marcus-etal-1993-building} with the standard splits (2-21 for training, 22 for validation, 23 for test) and the same preprocessing as in recent work \cite{kim-etal-2019-compound}, where we discard punctuation and lower-case all tokens. 
To explore the universality of the model across languages, we also run experiments on Chinese Penn Treebank (CTB) 8 \cite{10.1017/S135132490400364X}, on which we also remove punctuation. Note that in all settings, the training is conducted entirely on raw unannotated text.

\subsubsection{Results and Discussion}

\begin{table}
\begin{center}
\setlength{\tabcolsep}{3.pt}
\resizebox{0.48\textwidth}{!}{
\begin{tabular}{@{}c|l|c c|c@{}}
                    &       & \multicolumn{2}{c|}{WSJ}                      & \multicolumn{1}{c}{CTB}                         \\
Model               & cplx  &   $F_1$(M) & $F_1$ & $F_1$\\ \hline \hline
Left Branching (W)  & $O(n)$& - & 8.15  & 11.28 \\
Right Branching (W) & $O(n)$& - & 39.62 & 27.53 \\
Random Trees (W)    & $O(n)$& - & 17.76 & 20.17 \\
\hline
\bert-MASK (WP)      & $O(n^4)$&- & 37.39 & 33.24      \\
ON-LSTM (W)         & $O(n)$  &50.0\dag & 47.72 & 24.73 \\
DIORA (W)           & $O(n^3)$&58.9\dag & 51.42 & -  \\
C-PCFG (W)          & $O(n^3)$&\textbf{60.1}\dag & \textbf{54.08} & \textbf{49.95} \\ \hline
Ours (WP)           & $O(n)$  & - & 48.11 & 44.85                     \\ \hline \hline
DIORA (WP)          & $O(n^3)$& - & 43.94  & -  \\
C-PCFG (WP)         & $O(n^3)$&- & 49.76 &  60.34 \\
Ours (WP)           & $O(n)$  &- & \textbf{52.28} & \textbf{63.94} \\ \hline
\end{tabular}
}
\end{center}
\caption{Unsupervised parsing results with word (W) or word-piece (WP) as input. 
        Values with $^{\dag}$ are taken from \newcite{kim-etal-2019-compound}. 
        $F_1$(M) describes the max.\ score of 4 runs with different random seeds. 
        The $F_1$ column shows results of our runs with a random seed.
        The bottom three systems take word-pieces as input, and are also measured against word-piece level golden trees.}
\label{tbl:constituency_parsing}
\end{table}

Table~\ref{tbl:constituency_parsing} provides the unlabeled $F_1$ scores of all systems on the WSJ and CTB test sets. 
It is clear that all systems perform better than left/right branching and random trees.
Word-level C-PCFG (W) performs best on both the WSJ and CTB test sets when measuring against word-level gold standard trees. 
Our system performs better than ON-LSTM (W), but worse than DIORA (W) and C-PCFG (W). Still, this is a remarkable result.
Note that models such as C-PCFG are specially designed for unsupervised parsing, e.g., adopting 30 nonterminals, 60 preterminals, and a training objective that is well-aligned with 
unsupervised parsing.
In contrast, the objective of our model is that of bi-directional language modeling, and the derived binary
trees are merely a by-product of our model that happen to emerge naturally from the model's preference for structures that are conducive to better language modeling.

Another factor is the mismatch between our training and evaluation, where  
we train our model at the word-piece level, but evaluate against word-level gold trees. For comparison, we thus also considered DIORA (WP), C-PCFG (WP), and our system all trained on word-piece 
inputs, and evaluated against word-piece level gold trees. 
The last three lines show the results, with our system achieving the best $F_1$. As breaking words into word-pieces introduces word boundaries as new spans, while word boundaries are easier to recognize, the overall $F_1$ score may increase, especially on Chinese.

\begin{table}
\begin{center}
\setlength{\tabcolsep}{3.5pt}
\resizebox{0.48\textwidth}{!}{ 
\begin{tabular}{@{}cc| c c c c c c@{}}
 & (WP)  & WD & NNP & NP & VP & SBAR\\\hline \hline
\multirow{3}{*}{\rotatebox[origin=c]{90}{WSJ}} & DIORA  & 81.65 & 77.83 & 71.24 & 17.30 & 22.16\\
& C-PCFG                  & 74.26 & 66.44 & 65.01 & 23.63 & \textbf{40.40}\\
& Ours     & \textbf{99.24} & \textbf{86.76} & \textbf{72.59} & \textbf{24.74} & 39.81\\ \hline \hline
\multirow{2}{*}{\rotatebox[origin=c]{90}{CTB}} & C-PCFG &89.34 & - & 46.74 & \textbf{39.53} & -\\
 & Ours & \textbf{97.16} & - & \textbf{61.26} & 37.90 & -\\ \hline
\end{tabular}
}
\end{center}
\caption{Recall of constituents and words at word-piece level. WD means word.}
\label{tbl:unsupervised_chunking}
\end{table}

\paragraph{Analysis.} 
In order to better understand why our model works better 
when evaluating on word-piece level golden trees, we compute the recall of constituents following
\newcite{DBLP:conf/naacl/KimRYKDM19} and \newcite{drozdov-etal-2020-unsupervised}.
Besides standard constituents, we also compare the recall of word-piece chunks and proper noun chunks. 
Proper noun chunks are extracted by finding adjacent unary nodes with same parent and tag NNP. 

Table~\ref{tbl:unsupervised_chunking} reports the recall scores for constituents and words on the WSJ and CTB test sets. 
Our model and DIORA perform better for small semantic units,
while C-PCFG better matches larger semantic units such as VP and SBAR. 
The recall of word chunks (WD) of our system is almost perfect and significantly better than for other algorithms.
Please note that all word-piece level models are trained fairly without using any boundary information.
Although it is trivial to recognize English word boundaries among word-pieces using rules, this is non-trivial for Chinese. 
Additionally, the recall of proper noun segments is as well significantly better for our model compared to other algorithms.

\begin{table*}[ht]
\begin{center}
\scalebox{0.95}{
\begin{tabular}{@{}c|l l l l|l l l l@{}}
        & \multicolumn{4}{c|}{WSJ}                      & \multicolumn{4}{c}{CTB}                         \\
Model   & \%$_\mathrm{all}$&\%$_{n\leq 10}$& \%$_{n\leq 20}$& \%$_{n\leq 40}$&\%$_{all}$&\%$_{n\leq 10}$&\%$_{n\leq 20}$&\%$_{n\leq 40}$    \\ \hline \hline
\bert-MASK (W)    & 53.53 & 77.03 & 55.46 & 44.66 & 48.56 & 68.89 & 47.27 & 36.62 \\
ON-LSTM (W)      & 61.43$^{\dag}$ & 77.05$^{\dag}$ & 62.99$^{\dag}$ & 55.94$^{\dag}$ & 36.48 & 58.57 & 34.08 & 26.59 \\
DIORA (W)        & 67.76 & 78.08 & 68.80 & 64.15 & ~~---~ & ~~---~ & ~~---~ & ~~---~\\
C-PCFG (W)       & \textbf{72.74}$^{\dag}$ & \textbf{85.10}$^{\dag}$ & \textbf{74.65}$^{\dag}$& \textbf{67.19}$^{\dag}$ & \textbf{64.41} & \textbf{75.54} & \textbf{65.89} & \textbf{58.16} \\ \hline\hline
DIORA (WP)       & 54.73 & 68.80 & 55.68 & 49.22 & ~~---~ & ~~---~ & ~~---~ & ~~---~ \\
C-PCFG (WP)      & 67.18 & \textbf{83.09} & 68.20 & 61.03 & 62.25 & \textbf{74.98} & 61.04 & 52.52 \\ \hline
Ours (WP)   & \textbf{69.29} & 80.29 & \textbf{70.29} & \textbf{64.79}& \textbf{64.74} & 74.42 & \textbf{63.86} & \textbf{59.20} \\ \hline
\end{tabular}
}
\end{center}
\caption{Compatibility with dependency trees. (W) denotes word level inputs, (WP) refers to word-piece level inputs. 
         \%$_\mathrm{all}$ denotes the accuracy on all test sentences, while \%$_{n \leq x}$ is the accuracy on sentences of up to $x$ words. 
         Values with $^{\dag}$ are evaluated with predicted trees from \newcite{kim-etal-2019-compound}}
\label{tbl:dependency_compatibility}
\end{table*}

\subsection{Dependency Tree Compatibility}
We compared examples of trees inferred by our model with the corresponding ground truth constituency trees (see Appendix), 
encountering reasonable structures that are different from the constituent structure posited by the manually defined gold trees.
Experimental results of previous work \cite{drozdov-etal-2020-unsupervised,kim-etal-2019-compound}
also show significant variance with different random seeds. 
Thus, we hypothesize that an isomorphy-focused $F_1$ evaluation with respect to gold constituency trees is insufficient to evaluate how reasonable the induced structures are. 
In contrast, dependency grammar encodes semantic and syntactic relations directly, and has the best interlingual phrasal cohesion properties~\cite{fox-2002-phrasal}.
Therefore, we introduce dependency compatibility as an additional metric and re-evaluate all system outputs.

\subsubsection{Setup}\label{sec:dp_setup}
\paragraph{Baselines and Data.}
As our approach is a word-piece level pretrained model, to enable a fair comparison, we train all models on word-pieces and learn models with the same settings as in the original papers.
Evaluation at the word-piece level reveals the model's ability to learn structure from a smaller granularity. 
In this section, we keep the word-level gold trees unchanged and
invoke Stanford CoreNLP \cite{manning2014stanford} to convert the WSJ and CTB into dependency trees.

\begin{figure}
    \centering
    \includegraphics[width=0.48\textwidth]{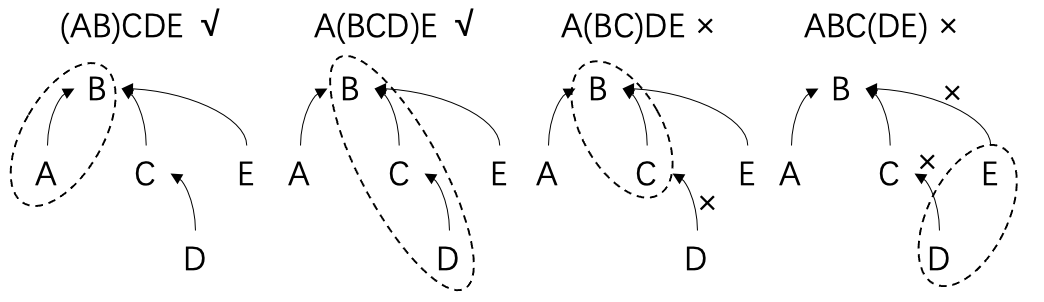}
    \caption{Examples of compatible and incompatible spans.}
    \label{fig:metric}
\end{figure}

\paragraph{Evaluation.}
Our metric is based on the notion of quantifying the compatibility of a tree by counting how many spans comply with dependency relations in the gold dependency tree.
Specifically, as illustrated in Figure \ref{fig:metric}, 
a span is deemed compatible with the ground truth if and only if this span forms an independent subtree.

Formally, given a gold dependency tree $\mathcal{D}$, we denote as $\mathcal{S}(\mathcal{D})$ the raw token sequence for $\mathcal{D}$.
Considering predicting a binary tree for word-level input, predicted spans in the binary tree are denoted as $\mathcal{Z}$.
For any span $z \in \mathcal{Z}$, the subgraph of $\mathcal{D}$ including nodes in $z$ and directional edges between them is referred to as $\mathcal{G}_{z}$. 
$\mathcal{O}(\mathcal{G}_{z})$ is defined as the set of nodes with parent nodes not in $\mathcal{G}_{z}$ and $\mathcal{I}(\mathcal{G}_{z})$ denotes the set of nodes whose child nodes are not in $\mathcal{G}_{z}$. 
Thus, $\left | \mathcal{O}(\mathcal{G}_{z}) \right |$ and $\left | \mathcal{I}(\mathcal{G}_{z}) \right |$ are the out-degree and in-degree of the subgraph $\mathcal{G}_{z}$.
Let $\mathrm{I}(z)$ denote whether $z$ is valid, defined as
\begin{equation}
\small
I(z)\left\{\begin{matrix}
1, & \left | \mathcal{O}(\mathcal{G}_{z})\right | = 1 \text{ and } \mathcal{I}(\mathcal{G}_{z}) \subseteq \mathcal{O}(\mathcal{G}_{z}) \\ 
0, & \text{otherwise.} 
\end{matrix}\right.
\end{equation}
For binary tree spans for word-piece level input, if $z$ breaks word-piece spans, then $\mathrm{I}(z)=0$. Otherwise, word-pieces are merged to words and the word-level logic is followed. Specifically, to make the results at the word and word-piece levels comparable, $\mathrm{I}(z)$ is forced to be zero if $z$ only covers a single word.
The final compatibility for $\mathcal{Z}$ is $\cfrac{\sum_{z \in \mathcal{Z}}^{}\mathrm{I}(z)}{|\mathcal{S}(\mathcal{D})| - 1}$.

\subsubsection{Results and Discussion}

Table \ref{tbl:dependency_compatibility} lists system results on the WSJ and CTB test sets.
\%$_\mathrm{all}$ refers to the accuracy on all test sentences, while \%$_{n \leq x}$ is the accuracy on sentences with up to $x$ words.
It is clear that the smaller granularity at the word-piece level makes this task harder.
Our model performs better than other systems at the word-piece level on both English and Chinese and even outperforms the baselines in many cases at the word level. 
It is worth noting that the result is evaluated on the same binary predicted trees as we use for unsupervised constituency parsing, yet 
our model outperforms baselines that perform better in Table~\ref{tbl:constituency_parsing}. 
One possible interpretation is that our approach learns to prefer structures different from human-defined phrase structure grammar
but self-consistent and compatible with a tree structure. 
To further understand the strengths and weaknesses of each baseline, 
we analyzed the compatibility of different sentence length ranges. 
Interestingly, we find that our approach performs better on long sentences compared with C-PCFG at the word-piece level. 
This shows that a bidirectional language modeling objective can learn to induce accurate structures even on very long sentences, on which custom-tailored methods may not work as well.

\section{Related Work}\label{sec:related_works}

\paragraph{Pre-trained models.}
Pre-trained models have achieved significant success across numerous tasks. 
ELMo~\cite{DBLP:conf/naacl/PetersNIGCLZ18}, pretrained on bidirectional language modeling based on bi-LSTMs, was the first model to show significant improvements across many downstream tasks. 
GPT~\cite{radford2018improving} replaces bi-LSTMs with a Transformer~\cite{vaswani2017attention}. 
As the global attention mechanism may reveal contextual information, 
it uses a left-to-right Transformer to predict the next word given the previous context. 
\bert~\cite{DBLP:conf/naacl/DevlinCLT19} proposes masked language modeling (MLM) to enable bidirectional modeling while avoiding contextual information leakage by directly masking part of input tokens.
As masking input tokens results in missing semantics, XLNET~\cite{yang2019xlnet} proposes permuted language modeling (PLM), 
where all bi-directional tokens are visible when predicting masked tokens. 
However, all aforementioned Transformer based models do not naturally capture positional information on their own and do not have explicit interpretable structural information,
which is an essential feature of natural language. To alleviate the above shortcomings, we extend pre-training and the Transformer model to structural language models.
\paragraph{Representation with structures.} 
In the line of work on learning a sentence representation with structures, 
\newcite{socher2011semi} proposed the first neural network model applying recursive autoencoders to learn sentence representations, but their approach constructs trees in a greedy way, 
and it is still unclear how autoencoders can perform against large pre-trained models (e.g., \bert).
\newcite{DBLP:conf/iclr/YogatamaBDGL17} jointly train their shift-reduce parser and sentence embedding components.
As their parser is not differentiable, they have to resort to reinforcement training,
but the learned structures collapse to trivial left/right branching trees.
The work of URNNG~\cite{DBLP:conf/naacl/KimRYKDM19} applies variational inference over latent trees to perform unsupervised optimization of the RNNG \cite{dyer-etal-2016-recurrent}, 
an RNN model that estimates a joint distribution over sentences and trees based on shift-reduce operations. \newcite{DBLP:journals/corr/MaillardCY17} propose an alternative approach, based on CKY parsing. The algorithm is made differentiable by using a soft-gating approach, which approximates discrete candidate selection by a probabilistic mixture of the constituents available in a given cell of the chart. This makes it possible to train with backpropagation. However, their model runs in $O(n^3)$ and they use Tree-LSTMs.

\section{Conclusion and Outlook}
In this paper, we have proposed an efficient CKY-based recursive Transformer to directly model hierarchical structure in linguistic utterances.
We have ascertained the effectiveness of our approach on language modeling and unsupervised parsing.
With the help of our efficient linear pruned tree induction algorithm, our model quickly learns interpretable tree structures without any syntactic supervision, which yet prove highly compatible with human-annotated trees. 
As future work, we are investigating pre-training our model on billion word corpora as done for \bert, and fine-tuning our model on downstream tasks.

\section*{Acknowledgements}

We thank Liqian Sun, the wife of Xiang Hu, for taking care of their newborn baby during critical phases, which enabled Xiang to polish the work and perform experiments.

\newpage
\bibliographystyle{acl_natbib} 
\bibliography{acl2021}

\onecolumn
\appendix
\section{Appendix: Tree Examples}

\begin{figure*}[!htb]
    \centering
    \includegraphics[width=0.98\textwidth]{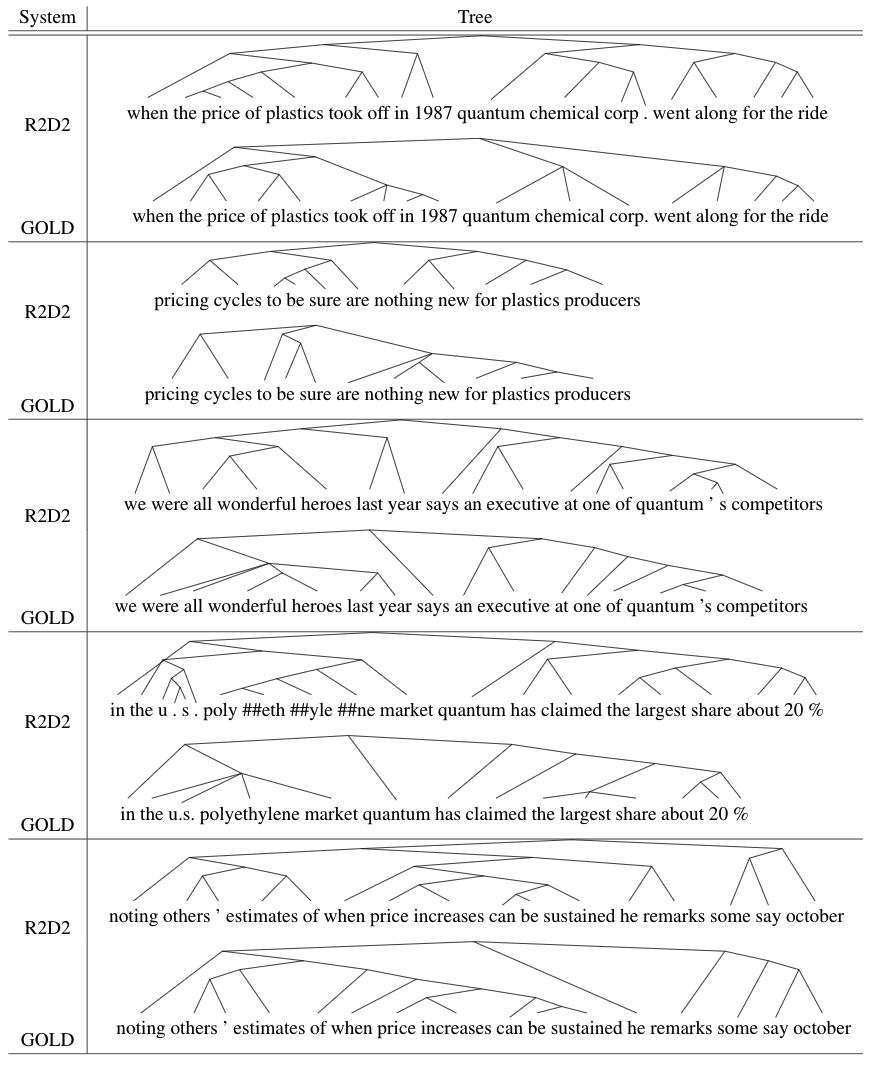}
\end{figure*}

\end{document}